\documentclass[runningheads]{llncs}

\usepackage[english]{babel}
\usepackage[utf8]{inputenc}
\usepackage[T1]{fontenc}
\usepackage{booktabs,multirow}
\usepackage[table]{xcolor}
\usepackage[colorinlistoftodos]{todonotes}
\usepackage{cite}
\usepackage{hyperref}
\usepackage{float} 
\begin{document}
%
\title{Text Classification Models for Form Entity Linking\thanks{This work was partially supported by Grant RTC-2017-6640-7; and by MCIN/AEI/10.13039/501100011033, under Grant PID2020-115225RB-I00.}}
\titlerunning{Text Classification Models for Form Entity Linking}


\author{María Villota\inst{1}\orcidID{0000-0002-1457-4270} \and C\'esar Dom\'inguez\inst{1}\orcidID{0000-0002-2081-7523} \and J\'onathan Heras\inst{1}\orcidID{0000-0003-4775-1306} \and Eloy Mata\inst{1}\orcidID{0000-0003-0538-4579} \and Vico Pascual\inst{1}\orcidID{0000-0003-3576-0889}}
\authorrunning{M. Villota et al.}
\institute{Department of Mathematics and Computer Science, University of La Rioja, Spain
\email{\{maria.villota, cesar.dominguez, jonathan.heras, eloy.mata, vico.pascual\}@unirioja.es}}

\maketitle

\begin{abstract}
Forms are a widespread type of template-based document used in a great variety of fields including, among others, administration, medicine, finance, or insurance. The automatic extraction of the information included in these documents is greatly demanded due to the increasing volume of forms that are generated in a daily basis. However, this is not a straightforward task when working with scanned forms because of the great diversity of templates with different location of form entities, and the quality of the scanned documents. In this context, there is a feature that is shared by all forms: they contain a collection of interlinked entities built as key-value (or label-value) pairs, together with other entities such as headers or images.  
In this work, we have tacked the problem of entity linking in forms by combining image processing techniques and a text classification model based on the BERT architecture. This approach achieves state-of-the-art results with a F1-score of 0.80 on the FUNSD dataset, a 5\% improvement regarding the best previous method. The code of this project is available at \url{https://github.com/mavillot/FUNSD-Entity-Linking}.

\end{abstract}

\section{Introduction}
Forms are template-based documents that contain a collection of interlinked entities built as key-value (also known as label-value or question-answer) pairs~\cite{docstruct}, together with other entities such as headers or images. These documents are used as a convenient way to collect and communicate data in lots of fields, including administration, medicine, finance, or insurance~\cite{Couasnon2014}. In these contexts, there is an enormous demand in digitising forms and extracting the data included in them~\cite{docstruct}; the latter is a task known as form understanding~\cite{funsd}. The form understanding task is especially challenging when working with scanned documents due to the diversity of templates, structures, layouts, and formats that can greatly vary among forms; and, also due to the different quality of the scanned document images~\cite{layoutlm}. 

Form understanding consists of two steps: \emph{form entity recognition} and \emph{form entity linking}~\cite{funsd}. In the former, the spatial layout and written information of forms are analysed to localise the position of form entities and to identify them as questions (keys), answers (values), or other entities present in the form. In the latter step, the extracted entities are interlinked to understand their relationships. Several approaches have been published in the literature in order to solve both tasks. Usually, they try to take advantage of both semantic text features and layout information of the forms by combining different methods~\cite{Carbonell, FUDGE, docstruct, MSAU-PAF, MTL-FoUn, SPADE}. In this work, we have focused on the problem of entity linking in forms using a new method that combines computer vision and natural language processing techniques. Namely, we can highlight the following contributions of this work:


\begin{itemize}
 \item We have proposed a new method for the task of entity linking in forms  that combines image processing techniques and a text classification model based on a transformer architecture. 
 \item For the text classification model, we have tested different architectures using transfer learning. The best model was obtained using the BERT architecture~\cite{bert}, which achieved a F1-score of 0.80; a 5\% improvement regarding the best previous method.
 \item Finally, we have publicly released all the code and models developed in this work  \url{https://github.com/mavillot/FUNSD-Entity-Linking}. 
 \end{itemize}

The rest of the paper is organised as follow. In the next section, we describe the FUNSD dataset~\cite{funsd} and the metrics used to test our approach for form entity linking. Subsequently, in Section~\ref{sec:related-work}, we conduct a literature review and analyse existing approaches to solve the task of form entity linking. The main features of our method are detailed in Section~\ref{sec:methods}, and a description of the obtained results is included in Section~\ref{sec:results}. This section also includes a comparison of our results with the results obtained by other methods on the FUNSD dataset. The paper ends with a conclusions and further work section.

\section{The FUNSD dataset}\label{sec:funsd}

Form understanding is a task that, up to now, has received little attention in the literature~\cite{funsd}. The scarcity of works in this area is mainly based on the absence of datasets of forms; such an absence was due to the sensitive information included in these documents. Therefore, a recent and important milestone in this area was the publication of the FUNSD dataset in 2019~\cite{funsd}, which was the first publicly available dataset that was developed with form understanding purposes. The FUNSD dataset is a fully annotated dataset of scanned forms from noisy and old documents. Moreover, baselines and metrics for the tasks of form entity recognition and form entity linking were provided.

The FUNSD Datatet~\cite{funsd} is freely available, and it contains 199 fully annotated images of forms that vary widely with respect to their structure and appearance. The forms come from different fields, e.g., marketing, advertising, and scientific reports; and they were sampled from the form type document of the RVL-CDIP dataset~\cite{icdar2015} which is composed of real grey scale images from the 1980s and 1990s. The documents have a quality with various types of noise added by successive scanning and printing procedures, and a low resolution (around 100 dpi). The annotation of each form is encoded in a JSON file. In the annotation files, each form is represented as a list of semantic entities that are interlinked. A semantic entity is described by a unique identifier, a label (chosen from four categories: question, answer, header, or other), a bounding box with the position of the entity, a list of words, and a list of links with the relationships among entities.  The 199 annotated forms contain more that 30,000 word-level annotations, 9,000 entities, and 5,000 relations. The dataset is split into 149 images in the training set and 50 in the testing set. 


%



In order to evaluate the tasks of form entity recognition and form entity linking in the FUNSD dataset, several metrics have been defined. We focus here on those related to the task of form entity linking that are F1-score, mean Average Precision (mAP) and mean Rank (mRank)~\cite{docstruct}. 
These metrics are based on the following values. The True Positive value (TP) is the number of links between a question and an answer that are correctly predicted. The False Positive (FP) value is the number of predicted links that do not exist. The False Negative (FN) value is the number of true links that are not found by the model. Finally, the number of unlinked answers and questions that are correctly predicted is the True Negative (TN) value. From these values the recall, precison and F1-score metrics are defined as follows:
$$\textit{recall}=\frac{TP}{TP+FN}, ~~\textit{precision}=\frac{TP}{TP+FP}, ~~ \textit{F1-score}=2 \cdot \frac{(precision \cdot recall)}{(precision + recall)}$$

In addition to the F1-score, other two metrics, called mAP and mRank, have been used in the literature to evaluate form entity linking algorithms. Both mAP and mRank are metrics used that come from the context of object detection and they can be interpreted as follows. The mAP value measures the average precision value for different recall values, so the larger the mAP value, the better; whereas, mRank measures the average number of wrong answers that rank higher than right answers, so the smaller mRank value, the better. More concretely, in order to define the {\it mAP value}, we need the confidence vector associated with each candidate question for each answer. Then, the AP value for each answer is defined as follows:
$$AP=\sum_{n}(R_n-R_{n-1})P_n$$
where $P_n$ and $R_n$ are the precision and recall at the $n^{th}$ threshold. Once we have calculated the AP for each answer, the mAP value is obtained by computing the mean value of all the APs values.

Finally, the {\it mRank metric} is defined as follows.  Given an answer $x$ that has $n$ candidate questions, $y_1,y_2,\dots,y_n$, and among them $m$ are correct (in FUNSD $m\leq 2$) with indices $i_1, i_2,\dots, i_m$ in an ascending order, the mRank metric is computed as follows. For the first answer $y_{i_1}$, the number of wrong answers that rank higher than it is $i_1-1$. For the second one, $y_{i_2}$, is $i_2-2$. Then, the Rank value is defined as: 
$$Rank= \sum_{k=1}^m i_k -k = \sum_{k=1}^m i_k - \frac{(1+m)m}{2}$$
and the mRank value is the mean value of the Rank values obtained from each answer.

\section{Related work}\label{sec:related-work}

Since its publication, the FUNSD dataset has been used as a benchmark by several works. With this dataset, the increased interest in automatic information extraction from semi-structured documents has been translated to form understanding with an increased number of publications~\cite{Carbonell, FUDGE, docstruct, MSAU-PAF, MTL-FoUn, SPADE}. In this section, we provide a summary of the different approaches existing in the literature to tackle the task of form entity linking using the FUNSD dataset.

The authors of the FUNSD dataset~\cite{funsd} not only provided such a dataset, but also developed a method for form entity linking. This method consisted in concatenating the entity feature representation  (extracted using the pretrained language model BERT~\cite{bert}) of every pair of entities in the form, and passing such a combined feature vector through a binary classifier constructed using a multi-layer perceptron. This baseline method achieved a F1-score of 0.04. From that seminal work, several methods for entity linking have been developed in the literature. 

In~\cite{docstruct}, a multimodal method to extract key-values pairs and build the hierarchical structure in documents for form entity linking was proposed. The form structure was considered as a tree-like hierarchy of text fragments, and the parent-child relation corresponded to key-value pairs in forms. The multimodal method included information from three sources: semantic features for the text fragments using a pre-trained language model (such as, BERT~\cite{bert} or RoBerta~\cite{roberta}), layout information showing the size and relative location of the text fragments, and visual aspects such as bold faces. In particular, given the hierarchical structure in the form, the superior counterpart for each text fragment was predicted.  This method was applied to the FUNSD dataset and it obtained a mAP of 0.72. An ablation study showed that the semantic and spatial features were the most valuable for completing the task~\cite{docstruct}. The combination of multiple aspects of forms was also used in~\cite{Bros}, where the authors created a model called BROS. This model added 2-D positional embeddings following 1-D BERT embeddings. This work also included a novel area-masking pre-training strategy designed for text blocks on the 2-D space and defined a graph-based decoder to capture the semantic relation between text blocks. The model was pre-trained on a great number of unlabelled documents in different domains. More concretely, they used the IIT-CDIP Test Collection~\cite{Lewis2006}, which contains more than 6 million scanned documents. Then, the model was fine-tuned using the embeddings on different contexts. In particular, the model was applied to form entity linking on the FUNSD dataset and it obtained a F1-score of 0.67. 

A different approach was proposed in~\cite{Carbonell}. In this work, a graph neural network was used to predict links between entities. In this approach, the document was represented as a graph whose nodes were the words previously provided by OCR, and the edges were initially created based on the distances of the top-left corner of the word bounding boxes. First, the model identified groups of words corresponding to entities by doing edge classification. Subsequently, the graph was used to perform entity labeling as a node classification and entity linking as edge classification tasks \cite{Carbonell}. This model obtained a F1-score of 0.39 in the entity linking task on the FUNSD dataset. On the contrary to the works~\cite{docstruct, Bros} that required a pre-training stage, this model only used FUNSD data to train it. A similar approach was used in~\cite{FUDGE} where a model named FUDGE using a graph neural network was proposed. In this work, the forms were also modelled as graphs where text segments were the vertices and pairwise text relationships were the edges. This model used a multi-step process involving text line detection, relationship proposals, graph editing to group text lines and prune edges, and finally relationship predictions~\cite{FUDGE}. In contrast to Carbonell et al.'s model~\cite{Carbonell}, FUDGE only used a purely visual solution (without including any language information), and it included an iterative method to allow the model to merge vertices and prune edges. Similarly to the work presented in~\cite{Carbonell}, FUDGE did not use any additional data from FUNSD in the training process. This models obtained a F1-score of 0.57 in the entity linking task. 

Finally, other three different approaches have been presented in the literature to solve the problem of form entity linking. First,  the authors of~\cite{SPADE} tackled the task of extracting information from a form as a dependency graph parsing problem. In particular, this work defines a model named SPADE. This model created a directed semantic relation graph of the tokens in the document using both linguistic and (two-dimensional) spatial information to parse the dependency~\cite{SPADE}. This model was trained to solve the entity linking problem on FUNSD dataset obtaining a F1-score of 0.41. Second, in~\cite{MTL-FoUn} a multi-task learning method was applied to form understanding. In this method, a model was defined to jointly learn three tasks simultaneously: word grouping, entity labelling, and entity linking. The idea was that the learning of the former two tasks (that were considered as auxiliary tasks) forced the model to learn to identify characteristics of the different entities types which, in turn helped the model to perform better in the entity linking task (the main task)~\cite{MTL-FoUn}. Using the LayoutLM model~\cite{layoutlm} as the backbone network for the three tasks, this work obtained a F1-score of 0.65. Finally, a hierarchical relationship extraction for form understanding was proposed in~\cite{MSAU-PAF}  to solve both entity labelling and entity linking tasks at the same time. The proposed model leveraged a specialised multi-stage encoder-decoder design, in conjunction with an efficient use of a self-attention mechanism and a box convolution to effectively capture the textual and spatial relations between 2D elements for solving entity labelling. Moreover, the method incorporated further techniques as part-intensity fields and part-association fields for solving entity linking. This last method was inspired by the techniques used in human pose estimation~\cite{MSAU-PAF}. This model was trained in FUNSD to solve together the classification and link prediction tasks, and it obtained in the entity linking problem a F1-score of 0.75.       



\section{Methods}\label{sec:methods}

\begin{figure}
    \centering
    \includegraphics[width=\linewidth]{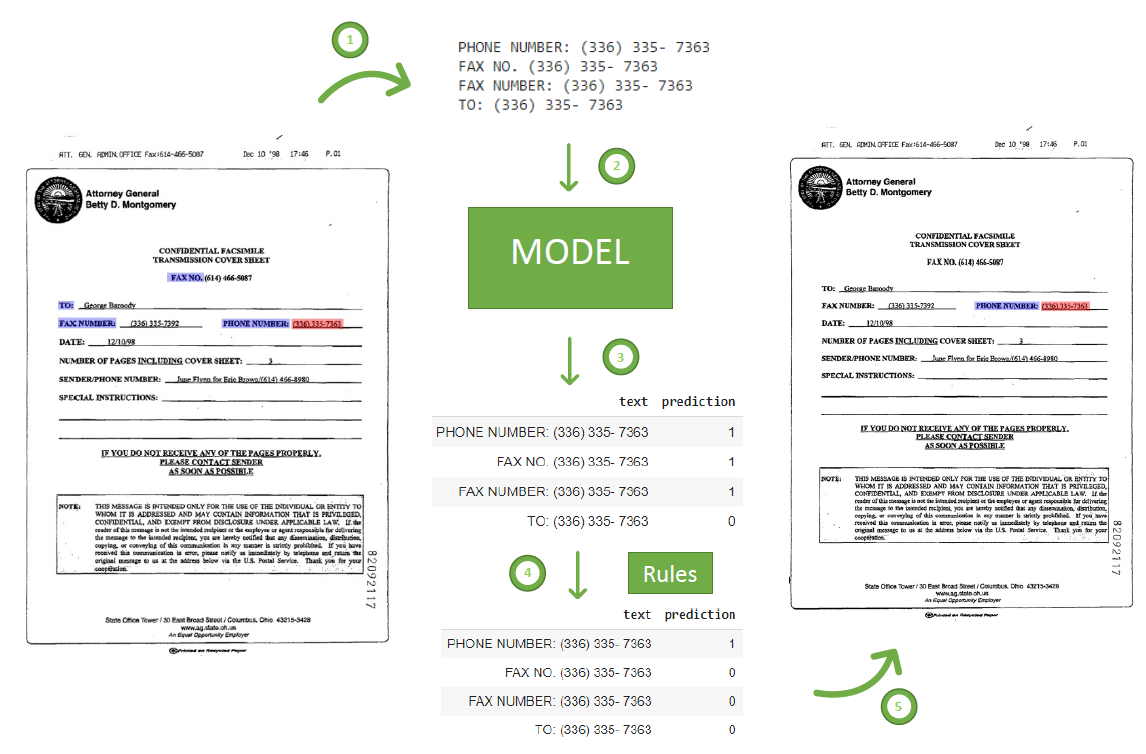}
    \caption{Pipeline of the proposed method. (1) From an answer, a set of candidate questions are identified. (2) Each combination of candidate question-answer is fed to a text classification model that (3) identifies the valid combinations of question-answer. (4) If more than one combination is valid, we apply a set of rules based on the distance between the question and the answer. (5) Finally, the results are returned.}
    \label{fig:pipeline}
\end{figure}

In this section, we present how we have combined image processing techniques and deep learning methods to perform the task of form entity linking, see Figure~\ref{fig:pipeline} for a summary of our method. For each answer that is found on a given form, we identify a set of candidate questions based on their distance to the answer; and, subsequently, we concatenate the text of each candidate question with the text of the answer, and use a text classification model to determine if that combination of question and answer makes sense. Finally, if multiple questions are valid for the given answer, we take the one that is closer to the answer. 

Now, we present how we have built the text classification model that determines if a combination of question and answer is valid. Text classification is a natural language processing task that consists in categorising a text into a set of predefined classes. Nowadays, this task is mainly tackled using deep learning models~\cite{bert}, and, namely, by training transformer-based architectures~\cite{attention}. Since it is not feasible to train this kind of model from scratch, due to the large amount of data that is required, we have applied transfer learning~\cite{Razavian14}, a technique that re-uses a model pretrained in a task where enough data is available. 

For our text classification models, we have fine-tuned several transformer-based language architectures; namely, BERT~\cite{bert}, DistilBert~\cite{distilbert}, Roberta~\cite{roberta}, DistilRoberta~\cite{distilbert}, and LayoutLM~\cite{layoutlm}. For fine-tuning the models, we replaced the head of each language model (that is, the last layer of the model), with a new head adapted to the binary classification task. Then, we trained the models for 6 epochs since more epochs produced overfitting. All the networks used in our experiments were implemented in Pytorch~\cite{pytorch}, and have been trained thanks to the functionality of the libraries Hugging Face~\cite{huggingface}, FastAI~\cite{fastai} and Blur~\cite{Blur} using the GPUs provided by the Google Colab environment~\cite{colab}, and using the by-default hyper-parameters for these models.

\section{Results}\label{sec:results}

In this section, we analyse the results achieved with our method. We start by exploring the performance of the studied text classification model, see Table~\ref{tab:comparison}. The best model for all the evaluated metrics is obtained using the BERT architecture. This model clearly overcomes the rest by a large margin, it achieves a F1-score of 0.80; whereas, the rest of the models obtain values lower than 0.70.

\begin{table}[]
    \centering
     \rowcolors{1}{white}{black!10!white}
    \begin{tabular}{lccc}
    \toprule
         & mAP & mRank & F1-score  \\
    \midrule
    BERT & \textbf{0.87} & \textbf{0.49} & \textbf{0.80}\\
    DistilBERT  & 0.79 & 0.79 & 0.68\\
    DistilRoBerta  & 0.76 & 0.95 & 0.65\\
    LayoutLM  & 0.79 & 0.81 & 0.69\\
    RoBerta  & 0.77 & 0.94 & 0.66\\
\bottomrule
    \end{tabular}
    \caption{Results achieved for entity linking by the tested text classification models. In bold face the best results.}
    \label{tab:comparison}
\end{table}

We additionally compare our proposed method with the existing algorithms available in the literature, see Table~\ref{tab:comparison2}. From such a comparison, we find that the performance of our method using the BERT model improves all the existing approaches. In addition, we can notice that our method, independently of the employed text classification model, obtains a better mAP and mRank than the algorithms available in the literature. This proves the effectiveness of combining image processing techniques and deep learning models in this context.

\begin{table}[]
    \centering
     \rowcolors{1}{white}{black!10!white}
    \begin{tabular}{lccc}
    \toprule
         & mAP & mRank & F1-score  \\
    \midrule
    BROS~\cite{Bros} & - & - & 0.67\\
    Carbonell et al.~\cite{Carbonell} & - & - & 0.39\\
    FUDGE~\cite{FUDGE} & - & - & 0.62 \\ 
FUNSD paper~\cite{funsd}& 0.23 & 11.68 & 0.04 \\
DocStruct Model~\cite{docstruct} & 0.72 & 2.89 & - \\
LayoutLM Word Level~\cite{MTL-FoUn} & 0.47 & 7.11 & -\\
MSAU-PAF~\cite{MSAU-PAF} & - & - & 0.75\\
MTL-FoUn~\cite{MTL-FoUn} & 0.71 & 1.32 & 0.65\\
Sequential Model~\cite{MTL-FoUn} & 0.65 & 1.45 & 0.61\\
SPADE~\cite{SPADE}& - & - & 0.41\\
\midrule
Ours  & \textbf{0.87} & \textbf{0.49} & \textbf{0.80}\\
\bottomrule
    \end{tabular}
    \caption{Comparison of our approach with existing methods for entity linking. In bold face the best results.}
    \label{tab:comparison2}
\end{table}

\section{Conclusion and Further work}

In this paper, we have proposed a method for form entity linking based on the combination of image processing techniques and text classification models. We tested several transformer-based models for text classification and reached the conclusion that a model based on the BERT architecture produces the best results. This approach has achieved state-of-the-art results for form entity linking in the FUNSD dataset, and shows the benefits of combining deep learning models with algorithms based on the existing knowledge about documents when working in contexts where annotated data is scarce.  

As further work, we are interested in applying our method to more recent documents since the FUNSD dataset is formed by old documents. The main challenges here are the privacy concerns raised when using form documents, and the issues related to the annotation of these documents, a time-consuming task that is instrumental to train and evaluate any deep learning model. Finally, we plan to adapt our approach to work with documents written on different languages.

\bibliographystyle{splncs04}
\bibliography{biblio}

\end{document}